\documentclass[letterpaper]{article}
\usepackage{aaai}
\usepackage{times}
\usepackage{helvet}
\usepackage{courier}
\usepackage{graphicx}
\usepackage{xspace}
\newcommand{\algo}{ComicGAN\xspace}
\begin{document}
 \author{Ben Proven-Bessel\textsuperscript{1},
Zilong Zhao\textsuperscript{2},
Lydia Chen\textsuperscript{3}\\
\textsuperscript{1,2,3}{Delft University of Technology}\\
\textsuperscript{1}{B.L.S.Provan-Bessell@student.tudelft.nl},
\textsuperscript{2}{Z.Zhao-8@tudelft.nl}, 
\textsuperscript{3}{lydiaychen@ieee.org}}


%
\title{\algo: Text-to-Comic Generative Adversarial Network}
\maketitle
\begin{abstract}
Drawing and annotating comic illustrations is a complex and difficult process. No existing machine learning algorithms have been developed to create comic illustrations based on descriptions of illustrations, or the dialogue in comics. Moreover, it is not known if a generative adversarial network (GAN) can generate original comics that correspond to the dialogue and/or descriptions. GANs are successful in producing photo-realistic images, but this technology does not necessarily translate to generation of flawless comics. What is more, comic evaluation is a prominent challenge as common metrics such as Inception Score will not perform comparably, as they are designed to work on photos.
In this paper:
1. We implement \algo, a novel text-to-comic  pipeline based on a text-to-image GAN that synthesizes comics according to text descriptions. 2. We describe an in-depth empirical study of the technical difficulties of comic generation using GAN’s.
\algo has two novel features: \textit{(i)} text description creation from labels via permutation and augmentation, and \textit{(ii)} custom image encoding with Convolutional Neural Networks. We extensively evaluate the proposed \algo in two scenarios, namely image generation from descriptions, and image generation from dialogue. Our results on 1000 Dilbert comic panels and 6000 descriptions show synthetic comic panels from text inputs resemble original \textit{Dilbert} panels. Novel methods for text description creation and custom image encoding brought improvements to Frechet Inception Distance, detail, and overall image quality over baseline algorithms. Generating illustrations from descriptions provided clear comics including characters and colours that were specified in the descriptions.
\end{abstract}

\section{Introduction}
\begin{figure}[t]
     \centering
         \includegraphics[width=0.15\textwidth]{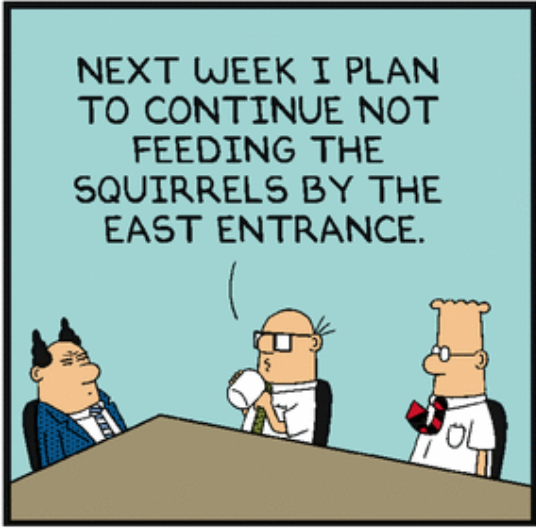}
         \includegraphics[width=0.148\textwidth]{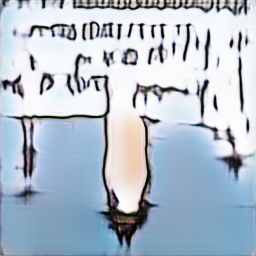}
         \includegraphics[width=0.15\textwidth]{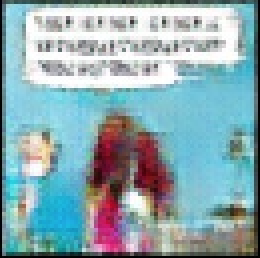}
     \caption{Example comic panels generated by standard 'out-of-the-box' versions of text-to-image and deep convolutional GAN respectively versus an original Dilbert comic panel from 2008/04/24 by Scott Adams. Unrecognisable, blurry characters and low image resolution should be improved upon.}
     \label{fig:intro-ex}
\end{figure}

Comics are amusing, fun, and bring a humorous aspect to our everyday lives. They join text and illustrations together in a unique and complex way.
Here we explore the possibility of using machine learning processes to generate comic illustrations from the dialogue present in comics, and from descriptions of illustrations. Generative Adversarial Networks (GAN) \cite{goodfellow2014generative} are successful at creating new data that emulates real data, and have been applied to table~\cite{zhao2021ctab,zhao2021fed,kunar2021dtgan}, audio~\cite{DBLP:conf/nips/KongKB20}, and image production \cite{reed2016generative}. Automatic image synthesis from text using GANs has been researched for realistic images taken by cameras but has not been applied to the domain of comic illustrations. We show that the success of GANs with photos does not translate well to generating high quality comics, and standard algorithms that generate superb images, do not generate superb comics (See Fig. \ref{fig:intro-ex}). Generation of cartoons using deep convolutional GAN's has been attempted, but comparisons of the best models has not been documented. Text-to-comic synthesis has not before been attempted. Comics are different in their makeup compared to regular photos, which brings specific challenges using GANs for comic generation, and to the evaluation of the success of this process. Therefore, we seek to further the area of GAN research within the realm of comics, to generate entertaining, engaging, and high quality comics.

\begin{figure}[t]
\centering
    \includegraphics[width=0.5\textwidth]{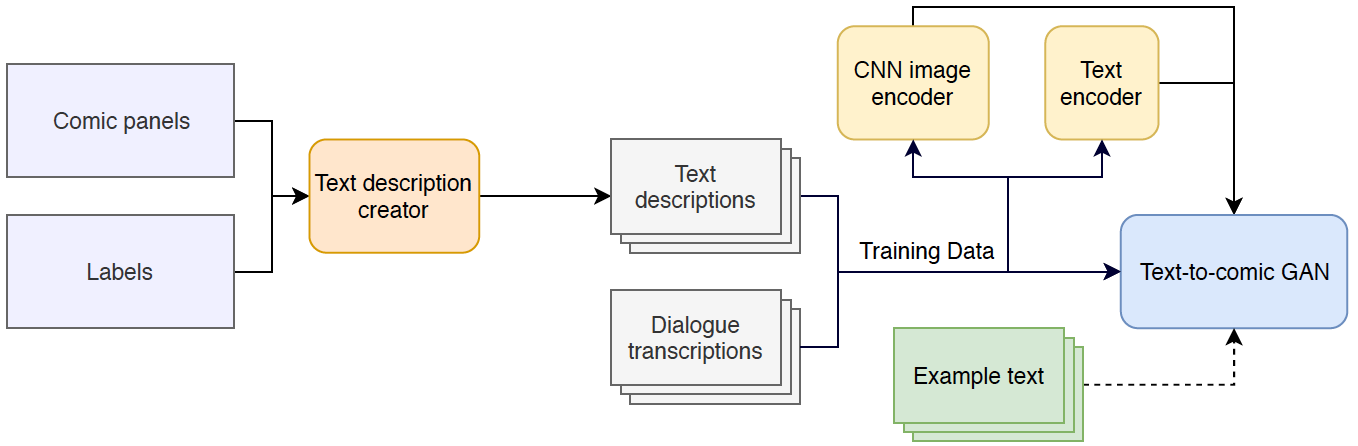}
    \caption{Text-to-comic model pipeline.}
    \label{fig:ttcdiag}
\end{figure}
The main goal of this research was to produce a model for text-to-comic generation. We apply our comic-generation methodology to \textit{Dilbert} \cite{dilbert}. The model should produce a comic based on a text input (see Fig. \ref{fig:ttcdiag}). To achieve this task, different research questions were investigated:
\begin{enumerate}
    \item What is unique about visual and textual features of comics that bring specific technical challenges to generate and evaluate comics?
    \item How to modify state-of-the-art text-to-image models to generate comics from text, including automatic text description creation and computer vision techniques for image encoding?
\end{enumerate}

Many challenges were evident from the start of this project. \textit{Text-to-image} models such as AttnGAN \cite{AttnGAN,MirrorGAN} learn semantic representation of the words within the image, e.g. that the word \textit{'red'} corresponds to the colour red within the image. The initial proposed use case of this model was to generate comics from the dialogue; however the connection between this text and the image is very abstract, it proved challenging to learn the text-to-image model relation. Therefore, descriptions with a direct connection to the images are required. Automatic processes to create multiple varied descriptions per image are necessary to satisfy the need for training data. Furthermore, methods in text-to-image algorithms are designed to work on photo-realistic images and do not necessarily translate optimally to work on comics. To overcome this issue, replacement modules designed specifically for comics were investigated to improve the quality of the generated panels.

A second major challenge involved evaluating the performance of the comic-generation pipeline. Comic generation from text is a task that has not yet been researched. There are no baseline models to compare to, and datasets with text descriptions for comics do not exist. Inception Score (IS), a metric commonly used to evaluate image quality is not applicable to evaluating comics, as it is trained for the purpose of photo realistic images. Text-to-image evaluation of how the image corresponds to the text is difficult, with metrics to measure this not commonly available to be applied to this project. Therefore extensive quantitative evaluation, and partial evaluation with Frechet Inception Distance (FID) \cite{fip} was necessary to judge results of the comics. The GAN models experimented with were very computationally expensive to train which limited the number of experiments in the project's timeframe.

The contributions of this research: 
\textit{(i)} The application of a \textit{text-to-image} model (AttnGAN) to create \algo: a new model to generate comic illustrations from text. This included implementation and investigation of: methods for generation of text descriptions of comics from labels and; custom visual feature extraction image encoder models.
\textit{(ii)} An in-depth study of technical challenges specific to generating comics such as: text correlation between comic and dialogue; extraction and uniqueness of visual features of comics and; evaluation of generated comics. Our results show that synthetic comic panels remarkably resemble original comics and achieve a high FID quantitatively. \algo can further produce comics that match a text description.

\section{Related work}
\label{section:related-work}

Artificial image synthesis refers to the task of creating an image, usually based on some constraint. This is an active area of research, and while some techniques using recurrent neural networks \cite{gregor2015draw} have been successful, GAN's are the most prominent and effective machine learning model to generate new images \cite{gansurvey,deepfakes}. We summarize the relevant related studies into sections regarding technologies of: Text-to-image GAN; Comic feature extraction with Convolutional Neural Networks (CNN).

There are successful text-to-image GAN's that can generate a detailed image based on a text description of the image. The original text-to-image GAN \cite{reed2016generative} is based on a model with a single generator and single discriminator such as \cite{dcgan}. The input text description is encoded, and a resulting image that corresponds to the semantic meaning of the sentence is generated. This correspondence is checked by the discriminator model. Models such as MirrorGAN \cite{MirrorGAN}, AttnGAN \cite{AttnGAN}, and StackGAN\cite{zhang2017stackgan}, extend \cite{reed2016generative} by creating additional generators that add detail at the level of each word in the description. Additionally, multiple discriminator networks are added to check that the image being generated is representative of the sentence, and the individual words in the sentence.

Encoding and performing feature extraction on the output image is an important sub-task. Both high-level features, and a final vector representation of the image are used in the attentional generation of the image, and for verifying that the image corresponds to the semantics of the text input. The image encoder used in AttnGAN which is based on the Inception v3 CNN \cite{inceptionv3}, pre-trained on ImageNet dataset \cite{imagenet}, should be substituted for a CNN more suited to encoding comics. Object detection for manga \cite{mangaCNN,mangaObDet,fastermanga} have used CNN based techniques to extract features from the comics. Convolutional architectures used for feature extraction and classification such as VGG and Inception v3 \cite{inceptionv3,vgg} are very powerful, but also are quite complex.

Generating images with GANs from text is a domain specific task, and out of the box solutions do not generate high quality comics. Moreover, specialized techniques are often needed to capture the distinct features and characteristics of comics. Modifications and improvements will need to be made to GAN and CNN models to improve their performance on comics. 

\section{\algo: Comics illustration synthesis}
\label{section:method}
This section will discuss the methods used: to extract and create text descriptions that correspond to comics; to augment and modify AttnGAN to create comics from text; and build a CNN image encoder for comics. Implementation details on each will be provided in respective sections.

\subsection{Architecture of AttnGAN }
\label{section:attngan-description}
AttnGAN \cite{AttnGAN} is a state-of-the-art GAN-based model that generates images from text, using images with multiple corresponding text captions as training data. This model was reproduced and tested on original text-image datasets \cite{CUB}. This model was used as a baseline for experimentation of generation of comics, and was fine tuned and augmented to improve results. AttnGAN uses multiple generators and discriminators, with encoded input text and encoded images. First, a Deep Attentional Multimodal Similarity Model is trained. This model is created by training a text encoder, and an image encoder. The text encoder is a Recurrent Neural Network that encodes the input text into a feature vector. The image encoder is a Convolutional Neural Network that encodes an image into feature vector. The DAMSM trains the image and text encoders to learn the semantic representation of the text description within ground truth images, e.g. that the word red in the input text represents the colour red within the image. The GAN is then trained. The first generator stage creates an image based on the entire encoded input text. Further generator stages use encodings of the most important words, and the results of the previous generator stage to upscale the image resolution, and add detail. Each generator stage output is judged by a discriminator. The DAMSM compares the final generated image to the text input, and resulting loss is used to improve generators. \algo first enables the generation of comics with AttnGAN by introducing a method to create text descriptions from a labeled image dataset. It then attempts to further improve generation of comics by creating a CNN image encoder designed specifically for comics.

\subsection{Comic Feature Extractor CNN}
\label{sec:comic-cnn}

\begin{figure}[!h]
    \centering
    \includegraphics[width=0.4\textwidth]{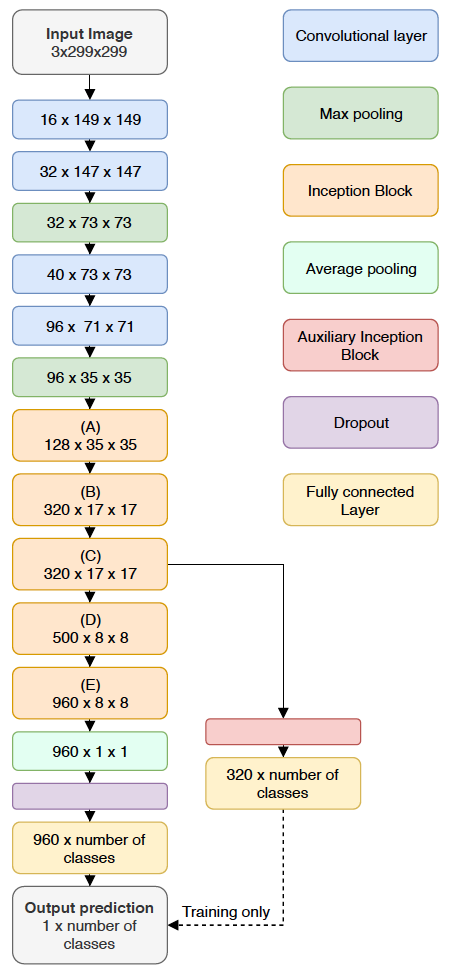}
    \caption{Comics CNN architecture. Sizes of inputs are indicated in each layer. The colours of each layer indicate the type of layer, as displayed by the key on the left. The input image is first processed by convolutional and max pooling layers, extracting basic features, and reducing spatial size but increasing depth. Mid-level, and final features are then extracted by a series of inception blocks, specifically designed to capture the different size of features present within an image, as the convolutional branches within each block contain varying stride lengths. The 17x17 mid-level regional features are flattened, and then used to predict the classification label with a final fully connected layer.}
    \label{fig:cnn_diag}
\end{figure}

To create a feature extractor for images, the Inception v3 CNN is trained on the ImageNet \cite{imagenet} dataset for multi-class classification. This same base network (with different final dense layers) is then used in AttnGAN as the feature extractor. Using the same transfer learning methodology, a multi-label classifier was trained specifically on comics, with the objective that a specialised CNN architecture could perform better feature extraction on comics, therefore creating higher quality comics with this input to the GAN. This Comics Classifier was trained and tested on the labels from which the comic descriptions were generated: the colour of the background and the characters in the comic. As it was time intensive and expensive to judge the results of image encoders while integrated as part of AttnGAN (due to the long training times), feature extraction capability was judged based on accuracy of the classifier. A higher accuracy means that the appropriate features are being extracted by the model.

The image encoder built into AttnGAN is based on the Inception v3 \cite{inceptionv3} CNN architecture. It is used to extract 17x17 regional features (which should represent individual features within an image such as a specific object or colour), and a final image encoding. To better capture these encodings of the features, a CNN created specifically to encode comics was designed and implemented. The architecture creates the same 17x17 regional features and final image encoding, but is adjusted and simplified to suit the simpler and more pronounced visual features of comics (compared to the more complex visual features of photo realistic images).

The comics CNN is constructed as follows: The input (image of size 299x299 pixels with 3 channels of red, green and blue) is put through two convolutional blocks, consisting of two 2-d convolutional layers and a max pooling layer, which reduce the spatial size from 299 to 35, while increasing the depth to 128.  An inception block
is created by creating three convolutional branches from the input. Each branch processes the input further with convolutional layers. Different size features are extracted as the layers of each have different stride lengths. Average or max pooling can be applied to the oupute of each branch, before the outputs are concatenated to give the output of the block. These blocks can replace convolutions with larger spatial features, while being less computationally expensive \cite{inceptionv3}. Three inception blocks increase depth, and reduce spatial size to 17. These blocks output 17x17 feature regions. This output is then processed through two more inception blocks, before final processing by average pooling, dropout, and flattening layers. A final fully connected layer outputs the final number of classes. As in inception v3, the comics CNN uses the feature regions while training to improve the loss of the model. The 17x17 features are used to predict the final class output as well. The final loss which is back-propagated through the model is: $loss + 0.4 \times feature loss$ Where $loss$ is the binary cross entropy loss between the final layer prediction and the actual output, and $feature loss$ is the binary cross entropy (BCE) loss combined with sigmoid activation ($\sigma$) (equation \ref{eq:bcelogits}, \ref{eq:ln}) between the feature layer prediction ($x$) and the actual output ($y$). The addition of the fractional loss $feature loss$ ensures that the regional features capture representative characteristics of the image.

\begin{equation}
    \label{eq:bcelogits}
    BCE loss = {l_1,...,l_n}^\top
\end{equation}

\begin{equation}
\label{eq:ln}
    l_n = -w_n[y_n \times log \sigma (x_n) + (1 - y_n)\times log(1 - \sigma(x_n))]
\end{equation}

\subsection{Text Description Formulation}
\label{section:text-form}
A text dataset corresponding to the comic panels was created. The steps and mechanisms used to label comics and subsequently create the text descriptions are described in this section.
\begin{enumerate}
    \item Comic panels were manually labelled with characters, and the background colour. Recurring characters had a unique number, while non-recurring characters were grouped together under one number. Background colour was chosen as the most dominant colour present.
    \item From these labels, descriptions were generated. To create a basic description, character numbers were replaced with the text names, and separated by \textit{"and"}. To complete the description, the background colour was added to the text in the form of \textit{"with colour background"}.
    \item To increase the number of descriptions per panel, augmentation was performed on the descriptions. The order of the characters could be permuted, providing $n factorial$ unique descriptions, where $n$ is the number of characters present. Descriptions were multiplied further by placing the background colour at the start, or end of the character portion of the description, $doubling$ the number of descriptions.
\end{enumerate}

Image synthesis from comics dialogue required the parsing of the comic dialogue. A resource for transcriptions of Dilbert comics \cite{arnold_2021} was downloaded, and a script was created to parse the dialogue into a dataset where each comic panel was associated with the text that appeared in it.


\section{Challenges of Comic Synthesis with GAN Technology}
\label{sec:tech-difficulties}
While GANs are competent at creating photo-realistic images, successfully applying this technology to generate comics has specific challenges and difficulties. This section describes an in-depth study of why text-to-image and GAN methods that work on images do not translate to comic-generation.

\subsection{Text Correlation between Comic and Dialogue}
\begin{figure}
\centering
    \includegraphics[width=0.5\textwidth]{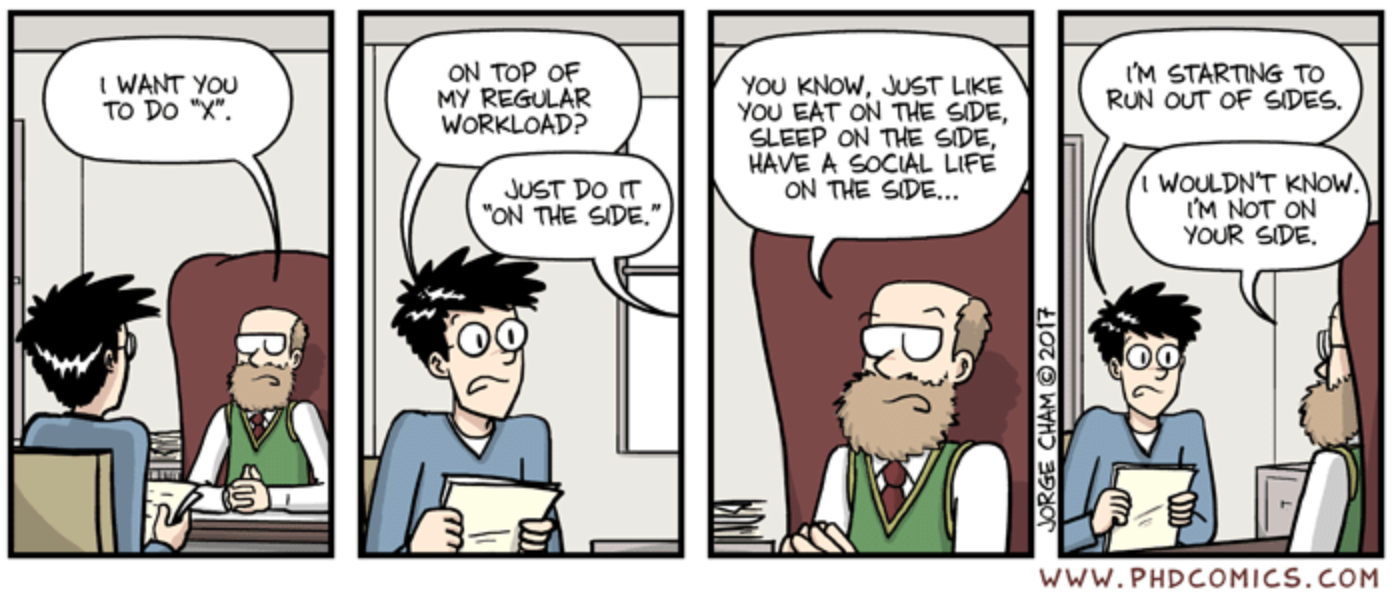}
    \caption{Example of \textit{PhD Comics} by Jorge Cham. The dialogue is not a clear description of what appears in the image.}
    \label{fig:comics}
\end{figure}

Initially it was thought that the comics dialogue could be used as input text for generating comics. However, problems with this approach were evident immediately. Text-to-image models are trained on images and text. Each image has corresponding textual descriptions which describe the subject/object within the image. There is a direct semantic connection between the text and the image. This is not at all what the comics dialogue and subtext is. There is no description of the characters or objects within the comic (see Fig. \ref{fig:comics}). The connection between the dialogue of comics and the illustration is abstract at best.

\begin{figure}[t]
     \centering

    \includegraphics[width=0.5\textwidth]{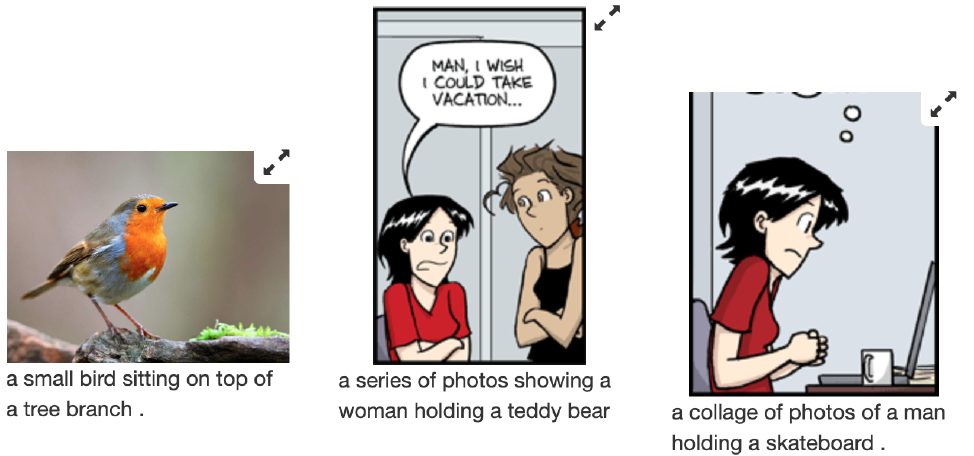}

     \caption{Automatically generated descriptions for comics and real life photos using IBM's image caption generator \cite{capgen}. Comics are not recognised properly.}
     \label{fig:auto-descriptions}
\end{figure}

\label{section:descriptionstrouble}
Automatic generation of descriptions was infeasible without the training of a custom ML model (see Fig. \ref{fig:auto-descriptions}). This is due to the fact that most description generators are trained on real life photographs, as opposed to illustrations. Therefore a study on correlation between comic and text had to be carried out. Images cannot be generated from text without a text dataset. Text descriptions had to be manually created. This led to questions: What type annotations should be created? What should they contain? How should they be composed such that the text-to-image models can learn the representation of the text that is present in the images?

A direct semantic connection between the text and contents of each image is needed for the text-to-image model to learn to create images that represent text.
Initial intuition led to emulation of the descriptions in other text-image datasets, which contained a description of the main subject of the image, using adjectives such as colours, shape, and size. This led to complicated descriptions, for which it was infeasible to create enough data to learn these descriptions (see Initial experiments Section for results). Final text datasets were created from labels of characters and background colour present in the comic (See section Text Description Formulation).

\subsection{Comic Image Feature Extraction and Uniqueness}
Comics are different in their makeup in comparison to real images. Computer vision applications trained specifically on photo realistic images are not necessarily effective when applied to comics. This was observed when descriptions generators did not work appropriately on comics, but did on real life images (see above section). 
Features of comics are different to those in images. Generally they are simpler and bolder: Black lines distinguish all characters and objects, characters each have unique hair, facial features and clothes. Backgrounds are often a single block colour, and do not contain the same level of variation and detail (see original Dilbert and PhD comics in figures \ref{fig:intro-ex}, \ref{fig:auto-descriptions}). 

\subsection{Evaluation of Generated Comics}
GANs are notoriously hard to evaluate. They  have  no  objective  loss  function, and results such as illustrations are subjective \cite{borji2019pros}. Due to the differences between photo images and comic illustrations, common evaluation metrics (e.g., IS) are not applicable for evaluating generated comics, as they are suitable only for photo-realistic images. Furthermore, Comic illustration synthesis from text has not been previously studied. Baseline algorithms are not available to provide comparisons on novel models created in this research.

\subsection{Comic Dataset Availability}
Absence of comic datasets with text descriptions proved to be a major problem. Text descriptions had to be manually created for each comic panel, which is time consuming and infeasible to carry out for the number of comic panels needed for a training set. Methods to create descriptions from character and colour labels helped partially automate text dataset creations, but it did not solve the problem. A set of comics, labelled with traits suitable to create descriptions was still needed. Such datasets do not exist. Due to time limitations of this project, only 2500 \textit{Dilbert} comics were labelled. Filtering of unusable comics, this reduced the dataset size to 1000 panels. Moreover, data-augmentation transformation applicable to images (e.g., brightness, rotation, cropping) are not applicable to comics, as they should have specific colours and orientations. Ideally, evaluation of GAN generation methods with more data, and multiple different comic datasets such as (\textit{Garfield, PhD comics}) is desired to form a more complete judgement. Furthermore, more descriptive text, rather than text derived from labels could improve generated image quality, and would serve as a good comparison to what text-to-image models can learn.



\section{Evaluation}
\label{section:evaluation}

\begin{figure*}[!h]
     \centering

         \includegraphics[width=\textwidth]{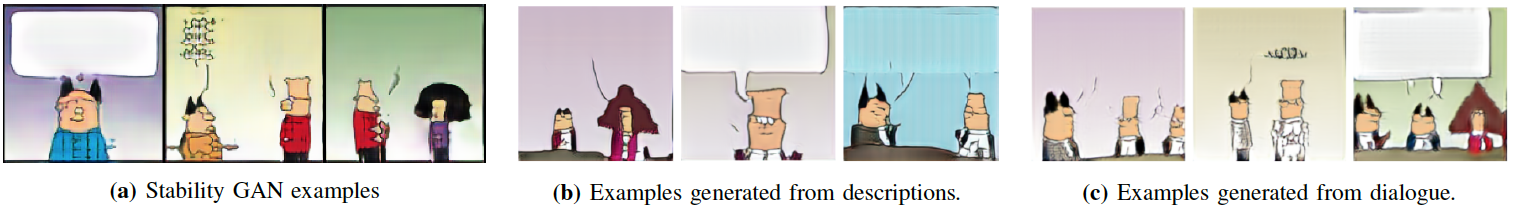}

        \caption{Generated comic panels. Models were trained on a Google-cloud virtual machine with a Nvidia K-80 GPU. PyTorch was the main implementation tool for models. Training for each model varied between 24-96 hour time periods.}
        \label{fig:overall-gen}
\end{figure*}
Over 15 experiments regarding different generative models were conducted. Comics CNN, and then comic generation with \algo are extensively evaluated.

\textbf{Image Datasets}.
\label{section:dataset}
\space \textit{Dilbert} was the main dataset used to evaluate the models created. Models were trained on single comic panels. Further image processing such as removing the text from the comics was done to simplify the dataset. Filtering of comics was performed based on labels of comics such as removal of uncommon and non recurring characters.

\textbf{Text Datasets}. \space
Dialogue datasets could be extracted automatically, for each comic that had a transcription. Descriptions were created manually, or translated from descriptions as described in Section \ref{section:text-form}.

Text datasets created:

\begin{itemize}
    \item Dialogue of the comic. The text that is inherent in the comics, spoken by the characters.
    \item Detailed character description. Text that describes the character. E.g. \textit{Dilbert is wearing a white shirt and a red tie and is sitting down by his computer.}
    \item General description of the comic. Text telling what characters are in the comic and what the background colour is. E.g. \textit{Dilbert and Alice with a green background.}
\end{itemize}

\textbf{Evaluation Metrics}.
\label{section:eval-metrics}
\space 

Fréchet inception distance (FID) \cite{fip} was used to quantitatively evaluate the quality of images generated by GAN's. FID compares the generated images to the images from which the GAN is created. Inception Score (IS) \cite{salimans2016improved} is another metric used to assess the quality of GAN generated images. FID improves over IS as it compares the features to the distribution of generated images $ N ( \mu_w, \Sigma_w) $ to the distribution of features of ground truth training images $ N ( \mu, \Sigma) $ (see equation \ref{eq:fid}), as opposed to IS which just evaluates the generated distribution. This is relevant when evaluating comics, as inception score is designed to judge photo-realistic images, not illustrations.

\begin{equation}
\label{eq:fid}
    FID = | \mu - \mu_w |^2 + tr(\Sigma + \Sigma_w - 2(\Sigma\Sigma_w)^\frac{1}{2})
\end{equation}

Generator and discriminator loss over epochs was used to evaluate the performance and stability of models, which can diagnose the vanishing gradient problem, and can give information as to whether the model has converged or will improve with more training.

Qualitative evaluation was also performed on models. Intermediate results at different checkpoints provide insight into in image generation. As images were being generated from text, the correspondence between the meaning of the text and the generated image was checked. This correspondence could be verified for text descriptions manually.

Multi-label classifiers were evaluated with accuracy, and F1 score. Accuracy is defined as the number of images for which the class predictions were correct, over the total number of predictions. In the context of multi-label classification, accuracy marks predictions that are partly correct as incorrect. Therefore F1 score defined as $2 \times \frac{precision \times recall}{precision + recall}$ was also used to give an indication of the proportion of classes in general predicted soundly.

\subsection{Multi-label Feature Extraction Analysis}
Three distinct experiments regarding CNN models for comic feature extraction were conducted. Training and validation accuracy over epochs, as well as accuracy and F1 score on a test set was used to analyse the performance of the multi-label classifier models.

Three different CNN architectures were experimented with:
\begin{itemize}
    \item Custom comics CNN architecture.
    \item Inception v3 pre-trained on ImageNet, final classification layer trained on comics.
    \item Inception v3 trained from scratch on comics.
\end{itemize}

\begin{table}[t]
    \centering
    \resizebox{\columnwidth}{!}{
    \begin{tabular}{ |c|c|c|c| } 
    \hline
    \textbf{CNN} & \textbf{Comics CNN} & \textbf{Inception pre-trained} & \textbf{Inception}\\ 
     \hline
     Accuracy (\%) & 76.9 & 80.77 & 71.1\\ 
     \hline
     F1-score & .936 & .945 & .917\\
     \hline
     Training time (minutes) & 37 & 52 & 156\\
     \hline
    \end{tabular}}
    \caption{CNN accuracy and F1 scores on test set, and training time to converge in minutes.}
    \label{tab:f1-acc-test}
\end{table}

\begin{figure}[t]
    \centering
         \includegraphics[width=0.23\textwidth]{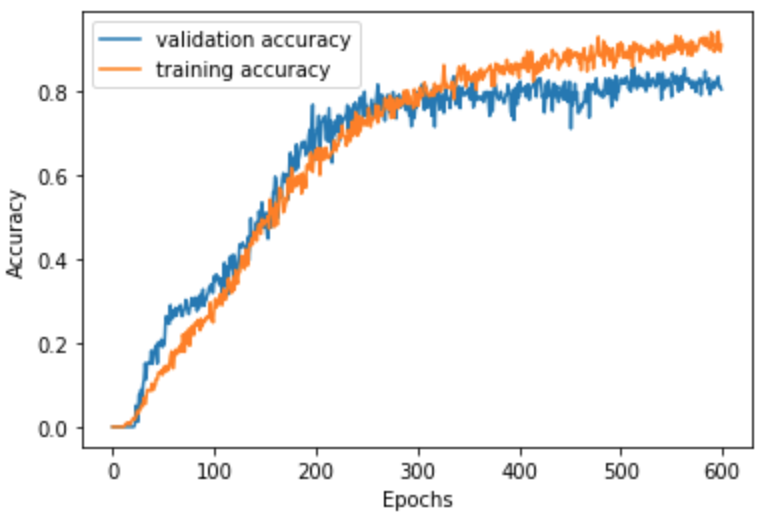}
         \includegraphics[width=0.23\textwidth]{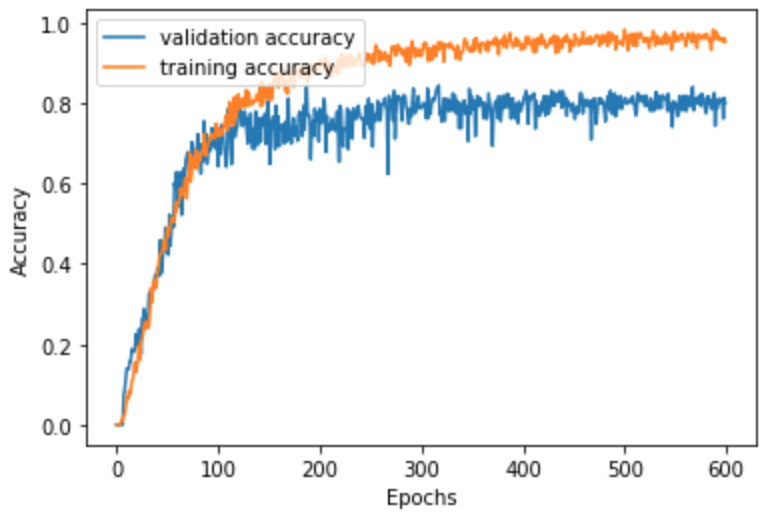}
    \caption{Training and validation accuracy over epochs for Inception v3 and Comics CNN multi-label classifiers trained from scratch.}
    \label{fig:train-val-graph}
\end{figure}

The inception v3 architecture pre-trained on imagenet performed best (see Section \ref{tab:f1-acc-test}, although it was prone to overfitting on the training set. Inception v3 architecture trained from scratch took far longer to train and did not reach the same accuracy compared to its pre-trained version. The comics CNN outperformed the scratch version of Inception, showing a 6\% improvement in accuracy on the test set, and taking far less time and epochs to converge (see Fig. \ref{fig:train-val-graph}). While the comics CNN was not able to exceed accuracy of the pre-trained version of inception, the improvement over the scratch version of inception indicates that the comics CNN is more suited to extracting features of comics. Therefore, a version of comics CNN pre-trained on a large dataset of comics should outperform the pre-trained version of inception.

\subsection{Text-to-Comic Analysis}
\label{section:tti-analysis}
In this section, text-to-image models based on AttnGAN and GAN generated comics are evaluated. FID of comics are compared, as well as extensive qualitative evaluation. First, \algo results will be compared to GAN generation of comics based on preliminary research. Then, 6 \algo are evaluated, based on 3 meaningful experiments: (1) Initial models using AttnGAN to create comics from text. (2) \algo trained on comic descriptions. (3) The \algo created from training data of comic dialogue only. Models for \algo used hyper-parameters based on that of the CUB bird model of AttnGAN, with slight changes to batch size. All parameters can be found in the configuration files in the source code.

\subsubsection{Overall Results Comparison}
\label{results}

This section will provide an overall comparison of results from the best generative models: DCGAN (StabilityGAN); Text-to-image GAN trained on descriptions (\algo); Text-to-image GAN trained on dialogue (\algo).

Fig. \ref{fig:overall-gen} shows generation results for the three GAN models. A preliminary empirical study on what Deep-Convolutional GAN (DCGAN) architecture was conducted, and StabilityGAN \cite{gan_stability} was found to be (quantitatively and qualitatively) the best DCGAN to generate comics without explicit conditions. The comics generated show a good variation in background colour, and they also capture characters recognisable as Dilbert, the pointy-haired boss, Alice and Wolly. StabilityGAN achieved the best FID score (see table \ref{tab:overall_fids}), with the two text-to-image models having comparable scores. The better FID scores correlated with longer training time and more data. While the StabilityGAN had the best FID score, and comics generated were more consistent, \algo generates higher resolution images (256x256 as opposed to 128x128), and you can control the characters that appear in the image, as well as the background colour.

\begin{table}[t]
    \centering
    \resizebox{\columnwidth}{!}
    {\begin{tabular}{ |c|c|c|c| } 
     \hline
     \textbf{Model} & \textbf{DCGAN} & \textbf{\algo (Description)} & \textbf{\algo (Dialogue)}\\ 
     \hline
     FID & 89.233 & 150.953 & 143.803 \\ 
     \hline
     Epochs & 1200 & 350 & 400 \\ 
     \hline
     
    \end{tabular}}
    \caption{Model FID and epochs trained}
    \label{tab:overall_fids}
\end{table}

A common downfall of all models was their inability to create characters such as Catbert and Dogbert. Their small size and infrequent appearances most likely contributing to the model's failure to reproduce them. This is a standard problem in machine learning when there is insufficient data.


\begin{table}[t]
    \centering
    \resizebox{\columnwidth}{!}{
    \begin{tabular}{ |c|c|c|c| }
     \hline
     \textbf{Model} & \textbf{Epoch 100} & \textbf{Epoch 200} & \textbf{Epoch 350}\\ 
     \hline
     FID & 200.258 & 182.163 & 150.953\\ 
     \hline
     
    \end{tabular}}
    \caption{Model FID score taken from epoch checkpoints}
    \label{tab:dilb3_fids}
\end{table}

\begin{figure}[t]
    \includegraphics[width=0.48\textwidth]{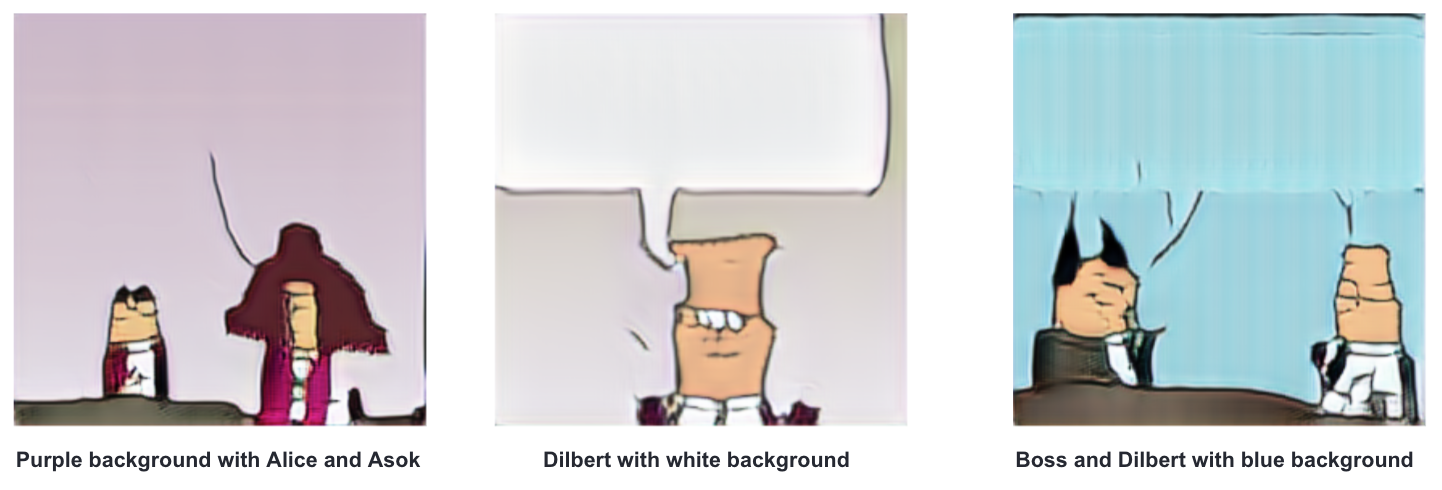}
  \hfill
  \caption{Generated examples comics. Text descriptions used to generate the illustrations appear below each respective comic.}
  \label{fig:gen3_examples}
\end{figure}

\begin{figure}[!t]
\centering
    \includegraphics[width=0.5\textwidth]{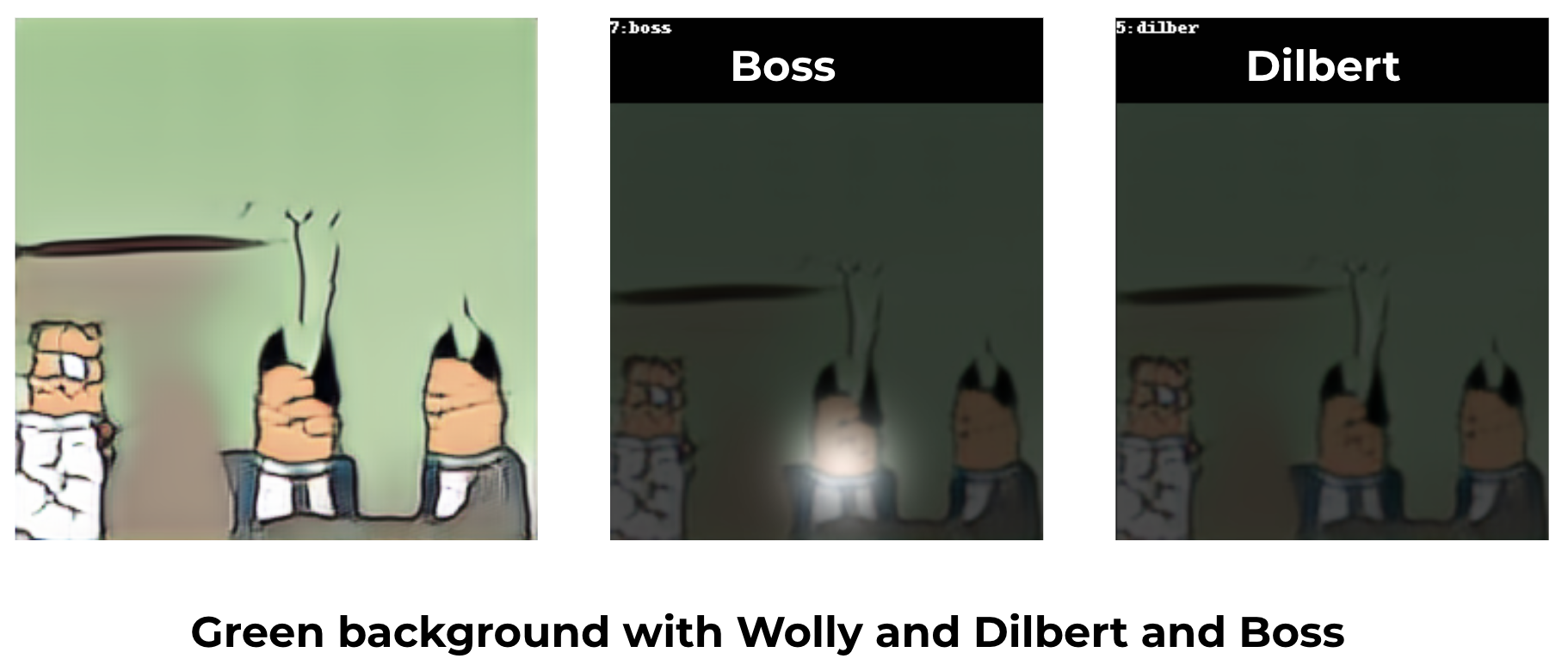}
    \caption{Attention mapping for words of \textit{Boss} and \textit{Dilbert}. As can be seen, the model learnt where the where the boss should be detailed, but not Dilbert.}
    \label{fig:gen3_attention}
\end{figure}
\subsubsection{Initial Experiments}
\label{section:initexp}
A text-to-image model trained on a small set of (75) \textit{Dilbert} images and corresponding descriptions was first created. This model was not trained for long enough, or on enough data, as results did not resemble any comic. A further text-to-image model trained on a dataset of 300 images of only Dilbert and/or Dogbert was created. Text descriptions were also limited to the form of \textit{Dilbert} or \textit{dog} or \textit{dilbert and dog}.

\begin{figure}[t]
    \centering
    \includegraphics[width=0.5\textwidth]{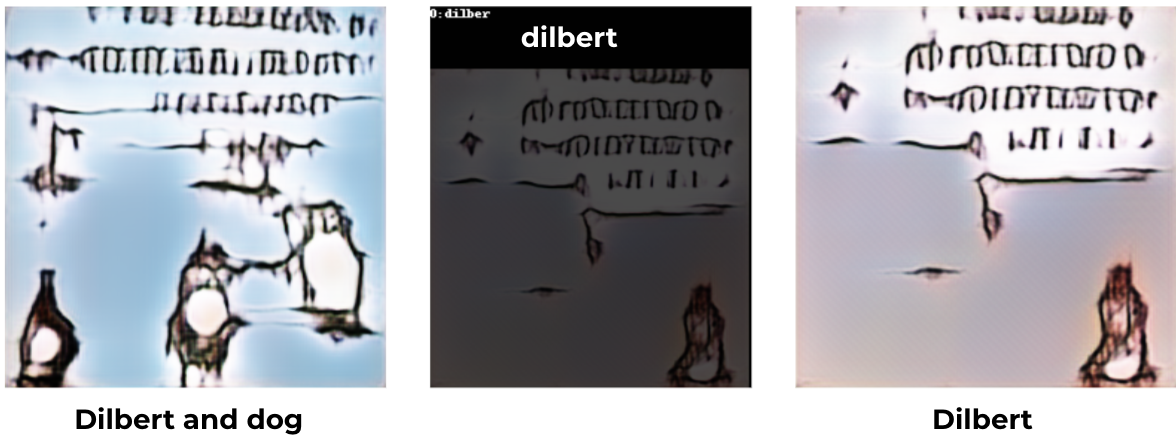}
    \caption{Images generated for captions of Dilbert and dog, and dog respectively. As can be seen in the centre attention map, the model was not able to learn the mapping of the word 'Dilbert' to its character.}
    \label{fig:gen_2_examples}
\end{figure}

While the results of this second model left much to be desired (See Fig. \ref{fig:gen_2_examples}), the improvement from the first model indicated that a larger dataset, with more descriptions per image would be necessary to generate clearer comics. Qualitative evaluation on a multiple generated images provided insight that the model had started to learn to draw characters and speech. These generator models were not possible to evaluate with FID, due to a minimum requirement of 2048 images in both training and generated images set. 

\subsubsection{Character and Colour Analysis}
A text-to-image model was trained on a dataset of descriptions of text and images created by methods described in \ref{section:text-form}. This formed a dataset of 1000 images, each with 6 corresponding character and colour descriptions, for 6000 text descriptions in total. Analysis for models saved at different points in training time was carried out, which was used to generate images for 2048 sample text descriptions.

Further training time improved the quality on the images. Characters became clearer, and the general quality of images got better. This is supported by the FID score results (see table \ref{tab:dilb3_fids}). Discriminator loss was generally much lower, with vanishing gradient starting to become a problem towards the end of training, which led to the conclusion that for the current data, the model was close to convergence.

Based on qualitative evaluation, this model learned the main features of the comics, and the text. Coloured backgrounds matched the text description, and the colour gradient was almost identical to that seen on the real \textit{Dilbert} comics. Almost all characters present in the training set could be recognised (see Fig. \ref{fig:gen3_attention}). Dilbert's unique face shape and glasses, The Pointy haired boss's hair, Alice's hair, Asok's face shape and complexion, and Wolly's stature (to some extent) were all recognisable in the generated comics.
This attempt had some issues. For example, Wolly was often generated as a character that looked like Dilbert. While the generated images did capture most of the characters, they did not always correspond directly to the description.

Improvements could be brought by creating a more balanced dataset, with more evenly distributed numbers of background colours, and characters. This would teach the GAN to learn to create these characters and colour as opposed to ignore them.

\begin{table}[t]
    \centering
    \resizebox{\columnwidth}{!}{
    \begin{tabular}{ |c|c|c|c|c| } 
     \hline
     \textbf{Model} & \textbf{Epoch 100} & \textbf{Epoch 200} & \textbf{Epoch 300} & \textbf{Epoch 400}\\ 
     \hline
     FID & 231.912 & 217.680 & 170.554 & 143.803\\ 
     \hline
     
    \end{tabular}}
    \caption{Model FID score taken from epoch checkpoints, trained on dialogue descriptions}
    \label{tab:dilb_trans_fids}
\end{table}

\begin{figure}[t]
    \centering
     
         \includegraphics[width=0.45\textwidth]{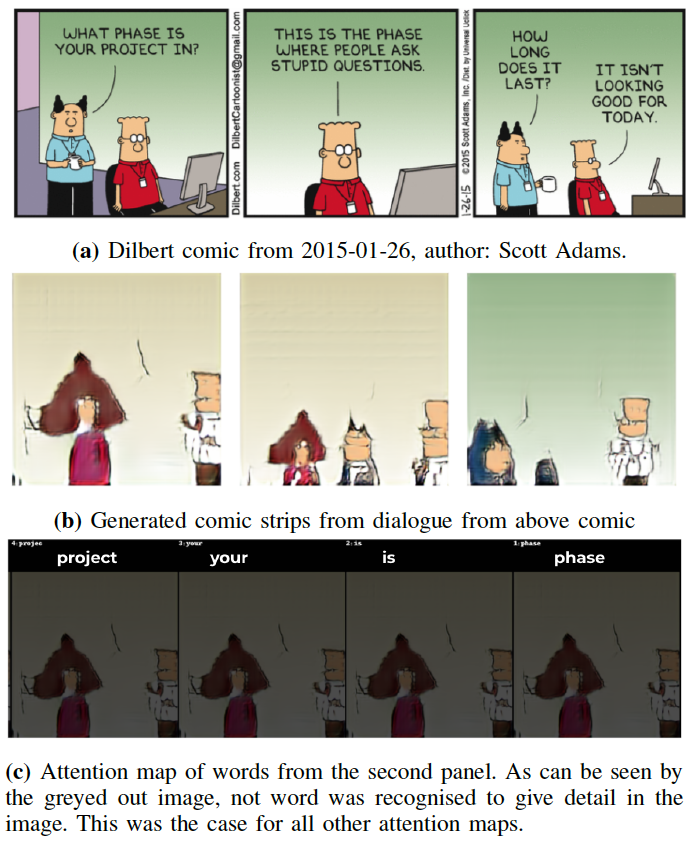}
    \caption{Ground truth comic, generated comic and attentional map for \algo trained on dialogue.}
    \label{fig:trans_gen_image}
\end{figure}

\subsubsection{Image Generation from Dialogue}
Generation of comics from dialogue only was experimented with. An extracted dataset of comic dialogue transcriptions was available \cite{arnold_2021}. This text was then paired to its respective comic panel. This formed a text-annotated image dataset. 2250 comic panels along with 2250 corresponding text dialogues were used to train the AttnGAN model. Images were generated on text used in the training set, and text not in the training set.

The GAN did learn to create comics that resembled the original comics (see Fig. \ref{fig:trans_gen_image} (a, b)). However the images generated had no correspondence to the dialogue text. This result can be seen in Fig. \ref{fig:trans_gen_image} (c), where the attention maps are not able to link input words to details in the image. This result is as expected. The model was able to learn to create comics with recognizable characters, but they lacked detail, and connection with the input text.

\section{Future Improvements and Suggested Work}
This section will describe work that could provide significant improvements to comic generation with GANs. These suggestions are based on findings of the empirical study, and experience gained generating comics with text-to-image GANs.

\subsection{Multi-Label Conditional GAN}
A multi-label conditional GAN could be created from a modified version of AttnGAN. While AttnGAN can produce high quality images from input text, it requires a large volume of detailed descriptions corresponding to images as training data. Multi-labeled image datasets are more commonly available and are easier to create by virtue of being simpler. AttnGAN uses a text-encoder (Recurrent Neural Network) to encode the input text as a feature vector. By simplifying the architecture of AttnGAN, label vectors could be passed in as input instead of text, foregoing the need for a text-encoder. Furthermore, the image-encoder could be simplified to a multi-label classifier, by reducing the output dimension of the CNN to that of the number of input classes with a linear layer. The DAMSM would compare the input label vector to the label vector predicted for the generated image. The 17x17 high-level regional features produced by the image-encoder would also have to be reshaped to the number of number of classes, and then compared to single encodings of the classes present in description. This modification of the model would remove the need to create the text-encoder, and remove any chance of misrepresentation or misinterpretation of the contents of the image that could be caused by the text encoder. This could improve image quality in two ways: images would have better correspondence to the input labels; images could be of better general quality. These improvements would occur because the DAMSM component could learn the labels and features on the images better, and therefore provide a more accurate loss to better train the generators. This modification could not be created due to time limitations of this research, but is proposed as future work.

Other improvements include: adapting StoryGAN \cite{li2019storygan} to the task of generating an entire comic strip, composed of multiple related panels that follow one after another; Further evaluation should include detailed analysis of various comic datasets, with multiple detailed text descriptions per image; Moreover, richer, more detailed text descriptions are necessary to fully test the capabilities of \algo. Due to the lack of annotated comic datasets, the models created performed very similarly to a multi-label GAN, which is not ideal for the technology in its' current state.

\section{Conclusions}
\label{section:conc}
The main purpose of this research  was to create a text-to-image GAN to create comics, and to apply the pipeline (using text input) to comics in the style of Dilbert. We have developed a text-to-image pipeline for comics synthesis based on existing methodology, and compared it to DCGAN comic generation. A novel text-to-image GAN, \algo was created by using automatic text description generation methods and custom comic feature extraction using CNN's.

Synthesis of comics illustrations was shown to be successful and representative of \textit{Dilbert} comics. StabilityGAN and \algo models brought significant improvements in FID score and image quality to comic generation over baseline models. A pipeline to create comic descriptions corresponding to comic panels to create text annotated comic datasets was designed and implemented. This enabled the training of text-to-image synthesis models, which produce comic panels characteristic of their description. High quality comics can be created which include the characters and colour in the description; however characters were misinterpreted occasionally. The comics CNN designed for specific comics feature extraction outperformed inception v3, with data suggesting that a large comics database would further improve results with pre-training. 
Differences between comic illustrations and photos were highlighted, and possible solutions to improve GAN generation results were proposed.

\bibliographystyle{aaai} 
\bibliography{reference.bib}

\end{document}